\documentclass[letterpaper, 10 pt, conference]{ieeeconf}
\makeatletter
\let\NAT@parse\undefined
\makeatother

\IEEEoverridecommandlockouts                              %

\usepackage[numbers,sort&compress]{natbib}

\usepackage{multirow}
\usepackage{multicol}
\usepackage[bookmarks=true]{hyperref}
\usepackage{xcolor}
\usepackage{color, colortbl}
\usepackage{tablefootnote} %
\usepackage{subcaption}
\usepackage{graphicx}
\usepackage{float}
\usepackage{amssymb}
\usepackage{amsmath}
\usepackage{tabularx}

\usepackage[capitalize]{cleveref}
\crefname{section}{Sec.}{Secs.}
\Crefname{section}{Section}{Sections}
\Crefname{table}{Table}{Tables}
\crefname{table}{Tab.}{Tabs.}

\usepackage{mathtools}

\def\eg{\emph{e.g.}}
\def\ie{\emph{i.e.}}

\usepackage{tikz}

\usepackage{pifont}%
\newcommand{\cmark}{\ding{51}}%
\newcommand{\xmark}{\ding{55}}%

\usepackage{soul}

\definecolor{Gray}{gray}{0.9}
\definecolor{LightGray}{gray}{0.7}
\newcolumntype{g}{>{\color{LightGray}}r}

\begin{document}
\bstctlcite{BSTcontrol}

\title{\LARGE \bf
StereoNavNet: Learning to Navigate using Stereo Cameras with Auxiliary Occupancy Voxels}

\author{Hongyu Li$^{1}$, Ta\c{s}k{\i}n Pad{\i}r$^2$, and Huaizu Jiang$^3$%
\thanks{This research is supported by the National Science Foundation under Award Number IIS-2310254.}
\thanks{$^{1}$Department of Computer Science, Brown University, Providence, RI. email: {\tt\small hongyu@brown.edu}}%
\thanks{$^{2}$Institute for Experiential Robotics, Northeastern University, Boston, MA. email: {\tt\small t.padir@northeastern.edu}}%
\thanks{$^{3}$Khoury College of Computer Sciences, Northeastern University, Boston, MA. email: {\tt\small h.jiang@northeastern.edu}}%
}

\maketitle

\newcommand{\lhy}[1]{\textcolor{red}{LHY: #1}}

\begin{abstract}
Visual navigation has received significant attention recently. Most of the prior works focus on predicting navigation actions based on semantic features extracted from visual encoders. However, these approaches often rely on large datasets and exhibit limited generalizability. In contrast, our approach draws inspiration from traditional navigation planners that operate on geometric representations, such as occupancy maps. We propose StereoNavNet (SNN), a novel visual navigation approach employing a modular learning framework comprising perception and policy modules. Within the perception module, we estimate an auxiliary 3D voxel occupancy grid from stereo RGB images and extract geometric features from it. These features, along with user-defined goals, are utilized by the policy module to predict navigation actions. Through extensive empirical evaluation, we demonstrate that SNN outperforms baseline approaches in terms of success rates, success weighted by path length, and navigation error. Furthermore, SNN exhibits better generalizability, characterized by maintaining leading performance when navigating across previously unseen environments.
\end{abstract}

\section{Introduction}
The intelligent navigation capability is essential for a robot to be integrated into our daily lives.
During the navigation, a robot needs to execute a sequence of actions to reach the desired goal, which may be a spatial coordinate, a specific object, or even a description in natural language~\cite{anderson_evaluation_2018, tellex_robots_2020}.
It is a challenging problem in robotics as the robot needs to move swiftly while effectively perceiving and avoiding unforeseeable obstacles.
Furthermore, the accurate perception and avoidance of obstacles often rely on data from expensive sensors.

\begin{figure}[ht!]
\centering
\includegraphics[width=\linewidth]{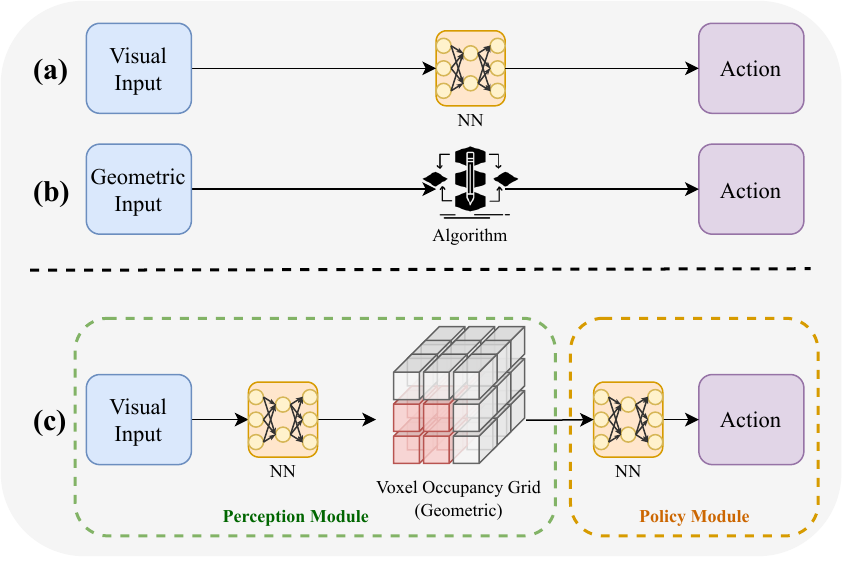}
\caption{\textbf{A high-level comparison.} Unlike the conventional visual navigation approach (a), where visual input is typically encoded into semantic features using a visual encoder for subsequent action prediction, we introduce a novel visual navigation network (c) inspired by traditional navigation approaches (b). We extract an auxiliary voxel occupancy grid from semantic features using a neural network (NN) and derive geometric obstacle features from it, which our policy network is conditioned on.
}
\label{fig: teaser}
\end{figure}

To extract scene semantics and improve affordability, visual navigation works focus on developing solutions that rely solely on low-cost vision sensors~\cite{gandhi_learning_2017, loquercio_dronet_2018, kendall_learning_2019, pan_agile_2019, kahn_land_2021, nguyen_motion_2022, tolani_visual_2021, shah_ving_2021}, for instance, RGB cameras. 
These approaches utilize deep learning-based visual encoders~\cite{he_deep_2016, yu_bisenet_2018, tan_efficientnet_2019, sandler_mobilenetv2_2018} to extract a feature vector from visual input,
which implicitly encapsulates semantic information such as obstacles and can be later used for obstacle avoidance. 
We refer to this class of methods as \emph{semantic} (\cref{fig: teaser} (a)). 
However, these semantic methods typically require large datasets or a long time for training and are susceptible to domain gaps when deployed in unseen environments.

Traditional navigation planners, such as Dynamic Window Approach (DWA)~\cite{fox_dynamic_1997} and Time Elastic Bands (TEB)~\cite{quinlan_elastic_1993, rosmann_kinodynamic_2017}, operate on geometric 2D/3D representations like a grid map or voxel occupancy grid. 
In these representations, free spaces and obstacles are clearly differentiated, often as zeros and ones in the 2D/3D grid (zeros mean free space). Therefore, we refer to this paradigm as \emph{geometric} (\cref{fig: teaser} (b)). 
Geometric planners exhibit appealing attributes compared with their semantic counterparts. 
For instance, such methods rarely suffer from the domain gap between various scene appearances as they do not directly take vision sensory data as input,
which is of particular interest in deploying an autonomous robot in practice. 
However, they have limited learning capacity and require accurate geometric information, which is often challenging to obtain during real-world deployment.

Can we combine the best of the two paradigms? 
Deep neural networks are suitable for processing raw sensory data, \eg, RGB images.
At the same time, a navigation policy learned from geometric scene representations, instead of visual input directly, may lead to higher learning efficiency and improved robustness to the domain gap~\cite{shen_situational_2019, wasserman_last-mile_2023}.
Several recent studies~\cite{kareer_vinl_2023, seo_learning_2023, wasserman_last-mile_2023, akmandor_deep_2022, wijmans_dd-ppo_2020, gervet_navigating_2023} have explored \textit{hybrid} approaches, leveraging both semantic and geometric features.
These approaches mostly follow the sim-to-real paradigm, relying on \textit{ground-truth depth maps} provided by simulated RGB-D sensors.
Nevertheless, \citet{bansal_combining_2020} and \citet{gervet_navigating_2023} observed that policies trained using this paradigm often encounter significant domain gaps in real-world deployment, primarily due to noisy depth measurements.

In this paper, we introduce StereoNavNet (SNN), a hybrid end-to-end visual navigation approach that encapsulates two modules: a perception module and a policy module, as shown in Fig.~\ref{fig: teaser} (c).
\textit{Without having access to any ground-truth depth information}, the perception module takes raw sensory data from stereo RGB cameras as input and outputs a voxel occupancy grid, which is a pure geometric representation of the scene.
It is then fed into the policy module, which produces a velocity command.

Such a \emph{modular} design endows our approach's desired properties for visual navigation.
First, it supports flexible training schedules.
We begin by training the perception and policy modules independently through modular learning~\cite{chaplot_object_2020, hu_modular_2023, gervet_navigating_2023, li_vihope_2023}.
Once both modules are trained to converge, we proceed to optimize the entire pipeline in an end-to-end fashion, which enables the backpropagation of perception errors to the policy end, compensating for the inevitable perception noises. 
Second, our policy module, conditioning only on geometric features, is more robust to the domain gap between training and testing data.

To confirm the effectiveness of our approach, we perform experiments in the OmniGibson framework~\cite{li_behavior-1k_2023} built on top of the NVIDIA Isaac Sim\textsuperscript{\texttrademark} simulator, which provides realistic rendering and physics simulation.
We compare against a semantic approach based on ResNet~\cite{he_deep_2016}, hybrid baselines using monocular and stereo depth estimation models~\cite{yang_depth_2024, shamsafar_mobilestereonet_2022}, and a geometric method using ground-truth depth information serving as the upper bound.
To evaluate our method, we collect an expert demonstration dataset for PointNav~\cite{anderson_evaluation_2018} task, consisting of 100 random trajectories across five environments (\cref{fig: train-scene}) using a privileged expert~\cite{chen_learning_2020, loquercio_learning_2021, sorokin_learning_2022} equipped with access to global ground-truth information.
We assess our method's robustness to the domain gap by validating it across seven novel environments. 
Our experiments show that our model achieves a higher success rate compared to the semantic approach and generalizes well in previously unseen environments, with a 35\% success rate in novel environments versus the semantic baseline's 21\%.

In summary, our main contributions are:
\begin{enumerate}
    \item We propose a novel visual navigation network encapsulating intermediate 3D geometric representation. Our approach is free of ground-truth depth map assumptions and relies purely on RGB images.
    \item We present extensive experiment results showing SNN outperforming other baseline methods in terms of success rates, success weighted by path length, and navigation error in both seen and novel environments.
\end{enumerate}

\section{Related Works}
There are two relevant fields to our paper: 1) visual navigation, the task we are solving, and 2) stereo vision, the approach we use to obtain the intermediate geometric representation.

\subsection{Visual Navigation}
\label{sec: related-works-nav}

Visual navigation is a type of navigation task that utilizes only visual input.
Depending on the various goals, different types of visual navigation tasks exist~\cite{savva_habitat_2019}.
For example, PointNav focuses on navigating to a spatial coordinate.
ObjectNav aims to find a type of object indoors, and ImageNav aims to locate the place where the goal image is taken.

We focus on the PointNav task in this paper since it is fundamental, \emph{i.e.} the obstacle avoidance ability required in this task is also needed by other tasks.
In this paper, we focus on extracting the geometry representation for obstacle avoidance. 
For other semantic navigation tasks, such as the ImageNav or the ObjectNav, it may require fusions with semantic information~\cite{gupta_cognitive_2017}.
Notably, \citet{gervet_navigating_2023} showed that semantic navigation tasks can be reduced to PointNav by proposing a module learning approach that transforms semantic goals into spatial goal coordinates.
 
Starting from the early works~\cite{gandhi_learning_2017, loquercio_dronet_2018, pan_agile_2019, zhao2e2}, a dominant paradigm is to predict actions using the semantic features extracted by a visual encoder~\cite{he_deep_2016, yu_bisenet_2018, tan_efficientnet_2019, sandler_mobilenetv2_2018}.
More recently, efforts have been directed towards refining learning strategies.
For instance, \citet{sorokin_learning_2022} achieve higher learning efficiency through privileged learning in an abstract world. 
\citet{kahn_land_2021} control the robot to avoid obstacles by minimizing the predicted human engagement probability. 
\citet{bansal_combining_2020} and \citet{tolani_visual_2021} predict waypoints and use an optimal controller to generate actions.
Nevertheless, relying solely on visual encoders necessitates large datasets and often results in limited generalizability.

Similar to ours, several works~\cite{kareer_vinl_2023, seo_learning_2023, wasserman_last-mile_2023, akmandor_deep_2022, wijmans_dd-ppo_2020, gervet_navigating_2023} also present visual navigation solutions that condition on geometric representations.
Notably, DD-PPO~\cite{wijmans_dd-ppo_2020} has achieved near-perfect performance on PointNav in Habitat-Sim benchmark~\cite{savva_habitat_2019}.
However, these works rely on the assumption of ground-truth depth information obtained by RGB-D sensors.
\citet{bansal_combining_2020} observed that the RGB-D approach has a significantly larger domain gap and worse performance than RGB counterparts due to noisy depth measurements in the real world.
Similarly, \citet{gervet_navigating_2023} found navigation errors in the real world are mainly caused by depth sensor errors.
This is primarily due to the limitations in the field of view, object size, material reflections, and lighting conditions~\cite{wang_detecting_2021, li2023stereovoxelnet, bansal_combining_2020, gervet_navigating_2023}. 
In light of these limitations, we assume no ground-truth depth information is provided and choose the stereo RGB sensor as the research platform.

\begin{figure*}[t!]
\vspace*{0.15cm}
\centering
\includegraphics[width=\linewidth]{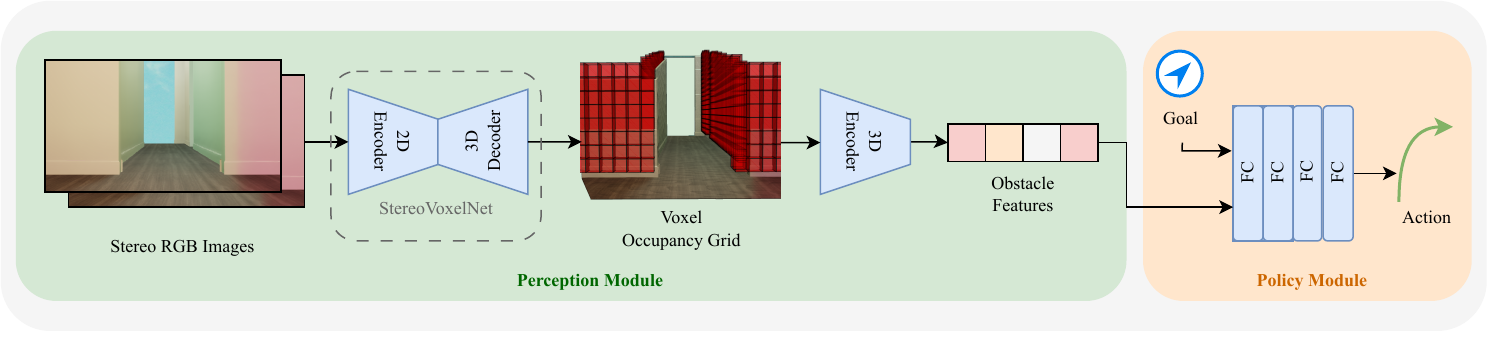}
\caption{\textbf{The network design of StereoNavNet.} We propose to extract the occupancy features from explicit geometry using the voxel occupancy grid and the curvature samples. The extracted features are used to predict an action using a four-layer MLP.
}
\label{fig: snn}
\end{figure*}

\subsection{Stereo Vision}
Stereo vision has been extensively used in robotics, enabling us to estimate depth by measuring the disparity between stereo images.
It relies on the principle of triangulation, mimicking the human visual system.
Classical methods like semi-global matching (SGM)~\cite{hirschmuller_accurate_2005} have been widely used in robot navigation~\cite{eppenberger_leveraging_2020, loquercio_learning_2021, seo_learning_2023}.
SGM aggregates matching costs over multiple directions and disparities and calculates the disparity for each pixel by minimizing a global energy function.
Even though these classical methods have the advantage in speed, they often demonstrate suboptimal performance compared with their learning-based counterparts~\cite{liu_local_2022, xu_attention_2022}. Furthermore, it is challenging to integrate the non-differentiable classical methods in an end-to-end learning framework.

Like many other tasks in computer vision, deep learning-based approaches have outperformed the classical ones by a large margin~\cite{zbontar_stereo_2016}.
Recent learning-based stereo models focus on building an end-to-end neural network that maps the stereo image pairs into disparity maps.
Inspired by traditional approaches, cost volume is commonly used to correlate the feature maps from the left and right images.
While the use of learning-based approaches is appealing, their high computational cost renders them impractical for real-time execution on an onboard computer during autonomous navigation~\cite{li2023stereovoxelnet}. 
To tackle this challenge, we introduced StereoVoxelNet~\cite{li2023stereovoxelnet} in the prior work, which offers an efficient deep-learning model for 3D perception. 
Instead of calculating the disparity map and transforming it into the 3D point cloud, StereoVoxelNet directly generates the 3D voxel occupancy grid from the stereo images. 
We employ this approach in our visual navigation framework for 3D perception, capitalizing on its efficiency and accuracy.

\section{Problem Formulation and Assumptions}
In this paper, we address the problem of PointNav \cite{anderson_evaluation_2018}, in which the robot operates in a static environment and has no prior maps or knowledge of the environment. 
In this problem, the locomotion problem is often abstracted out, \ie, there are no holes or stairs blocking movement, and the robot can movely freely to any location unless blocked by obstacles.
The robot relies solely on egocentric visual input from an onboard camera to perceive the environment. 
Given a user-defined point goal $g\in \mathbb{R}^3$, the agent generates a trajectory $\tau$ of actions $\tau = (a_1, a_2, \cdots, a_{|\tau|})$ to navigate the robot towards the goal without any collisions. 
$|\tau|$ represents the total steps taken to reach the goal.
Each action is a continuous velocity pair $a=(v, \omega)$, where $v$ denotes the linear velocity and $\omega$ denotes the angular velocity.
We also assume the ground-truth odometry is given, following the PointNav task setting~\cite{anderson_evaluation_2018}, so that the relative transformation between the robot and the goal is always accurate and known.

Since we focus on the sensing aspect in this paper instead of the navigation policy, we verify SNN on the simplest navigation agent, which is purely reactive~\cite{anderson_evaluation_2018} and Markovian.
Our method is model-free, which means that our policy module doesn't construct a map of the environment~\cite{gupta_cognitive_2017} nor store internal states across time steps using recurrent units~\cite{sak_long_2014}.

We assume the robot ``sees'' the world using stereo RGB cameras instead of RGB-D cameras, as discussed in \cref{sec: related-works-nav},
to mitigate potential navigation failures stemming from reflections, varying lighting conditions, etc., when transferred to real-world scenarios.
We choose the stereo setup as it provides more accurate 3D geometry than monocular~\cite{wang_pseudo-lidar_2019}.
In the experiment section, we provide numerical results on utilizing both monocular model~\cite{yang_depth_2024} and stereo model~\cite{shamsafar_mobilestereonet_2022} and demonstrate the efficacy of the stereo approach.

\section{StereoNavNet}
Our main contribution is the proposed StereoNavNet (SNN), which extracts an explicit 3D geometry representation from the visual observation and trains a policy network upon it.
SNN consists of two modules, the perception module and the policy module, as shown in Fig.~\ref{fig: snn}.
The perception module takes in stereo images as input and outputs occupancy voxels.
Subsequently, the occupancy feature and the user-defined goal $g$ are fed into the policy module (\cref{sec: policy_module}), which generates the velocity actions $a$.

\newcommand{\calG}{\mathcal{G}}

\subsection{Perception Module}
\label{sec: perception_module}
Our main motivation is to leverage a perception module $f_p$ to extract geometric features from the visual observation.
We hypothesize that a policy network trained on the geometric feature could lead to higher performance and a narrower domain gap since it does not directly take raw sensory data as input.
Specifically, the perception module aims to extract obstacle features $h_o$ from a pair of stereo images $(I_L, I_R)$, such that $h_o = f_p(I_L, I_R)$.
To achieve this goal, we first extract an intermediate geometry representation.
While there are various choices of 2D/3D representation, in this paper, we concentrate on utilizing the voxel occupancy grid $\calG$ as the 3D geometry representation.

\textbf{Extracting voxel occupancy grid.}
We define a robot-centric grid $\calG$,
\begin{equation}
    \calG = \{0,1\}^ {n_x \times n_y \times n_z}, n_x, n_y, n_z \in \mathbb{Z}^+.
\end{equation}
Here $n_x, n_y, n_z$ represents the number of voxels on $x,y,z$ axis, respectively, which point towards the left, top, and forward directions.
Each voxel is a cubic volume with side length $l_v$ and can be either occupied (denoted as 1) or unoccupied (denoted as 0). 
Therefore, the physical dimension of the voxel grid is a volume of size $(n_x\times l_v, n_y\times l_v, n_z\times l_v)$.

In order to estimate the voxel grid $\calG$, we employ our previous work, StereoVoxelNet~\cite{li2023stereovoxelnet}, which consists of a 2D-3D encoder-decoder structure. 
We refer readers to the original paper for more technical details.

The perception module is optimized through Intersection of Union (IoU) loss $\mathcal{L}_{p}$,
\begin{equation}
    \label{eqn: voxel_loss}
    \mathcal{L}_{p} = 1 - \frac{|\calG \cap \widehat{\calG}|}{|\calG \cup \widehat{\calG}|},
\end{equation}
where $\calG$ is the ground-truth voxel grid, and $\widehat{\calG}$ is the estimated voxel grid.

\textbf{Extracting geometric feature $h_o$.}
We extract the obstacle feature vector $h_o$ from the voxel occupancy grid $\calG$ using a set of four 3D convolutional layers with ReLU layers in between.
We set the size of $h_o$ to be 256 and configure the layers with channels of $[4,8,6,32]$ respectively, each with a kernel size of 4 and a stride of 2.

\begin{figure}[t!]
\vspace*{0.15cm}
\centering
    \begin{subfigure}{.48\columnwidth}
        \centering
        \includegraphics[width=\columnwidth]{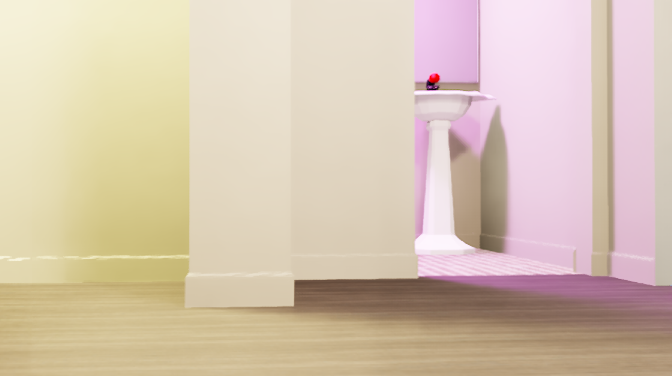}
    \end{subfigure}
    \begin{subfigure}{.48\columnwidth}
        \centering
        \includegraphics[width=\columnwidth]{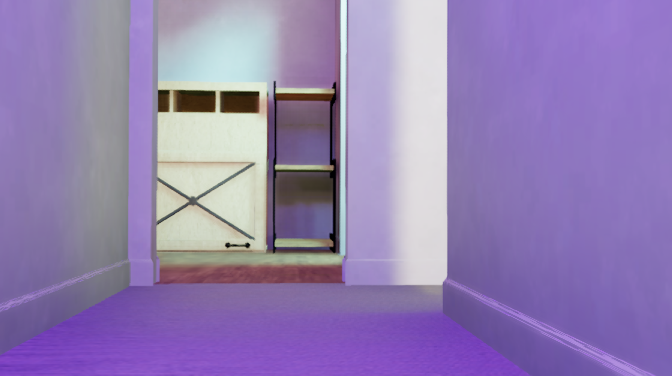}
    \end{subfigure}
    \begin{subfigure}{.48\columnwidth}
        \centering
        \includegraphics[width=\columnwidth]{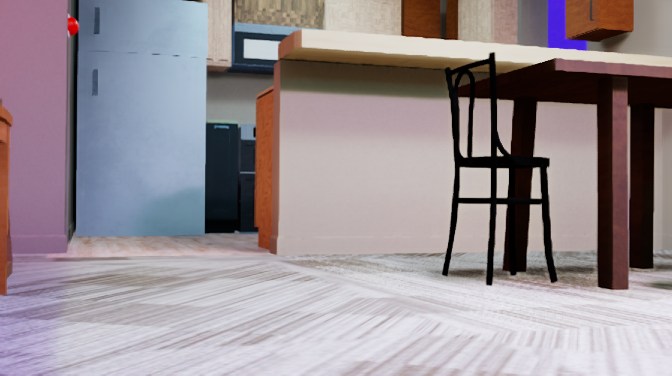}
    \end{subfigure}
    \begin{subfigure}{.48\columnwidth}
        \centering
        \includegraphics[width=\columnwidth]{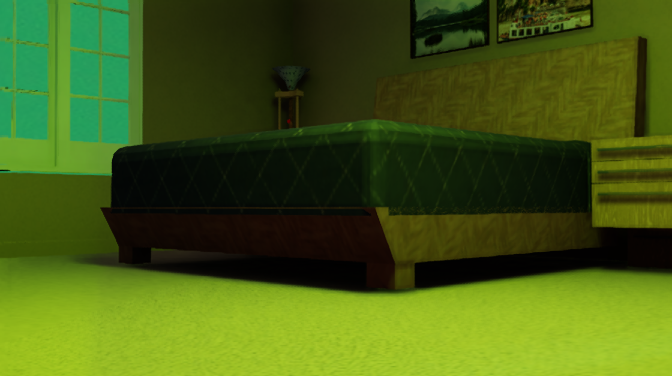}
    \end{subfigure}
    \begin{subfigure}{.48\columnwidth}
        \centering
        \includegraphics[width=\columnwidth]{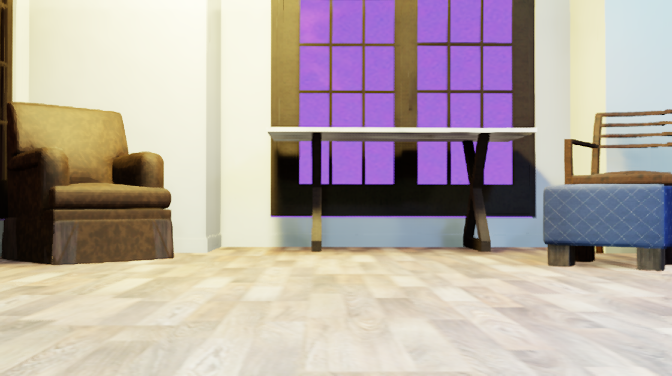}
    \end{subfigure}
    \begin{subfigure}{.48\columnwidth}
        \centering
        \includegraphics[width=\columnwidth]{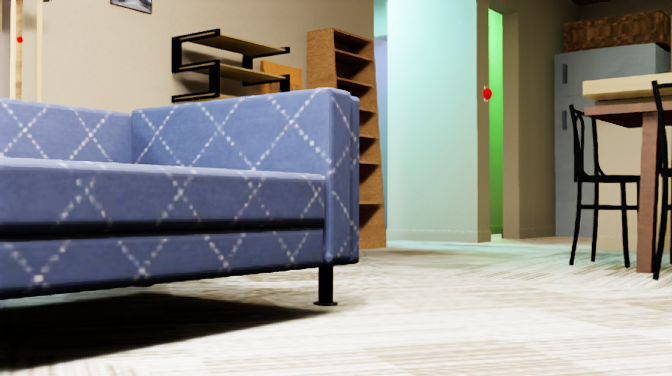}
    \end{subfigure}
\caption{\textbf{Training scenes}. We deploy a privileged agent in five scenes from OmniGibson~\cite{li_behavior-1k_2023} to collect an expert demonstration dataset.
}
\label{fig: train-scene}
\end{figure}

\begin{figure*}[ht!]
    \vspace*{0.15cm}
    \centering
    \includegraphics[width=\textwidth]{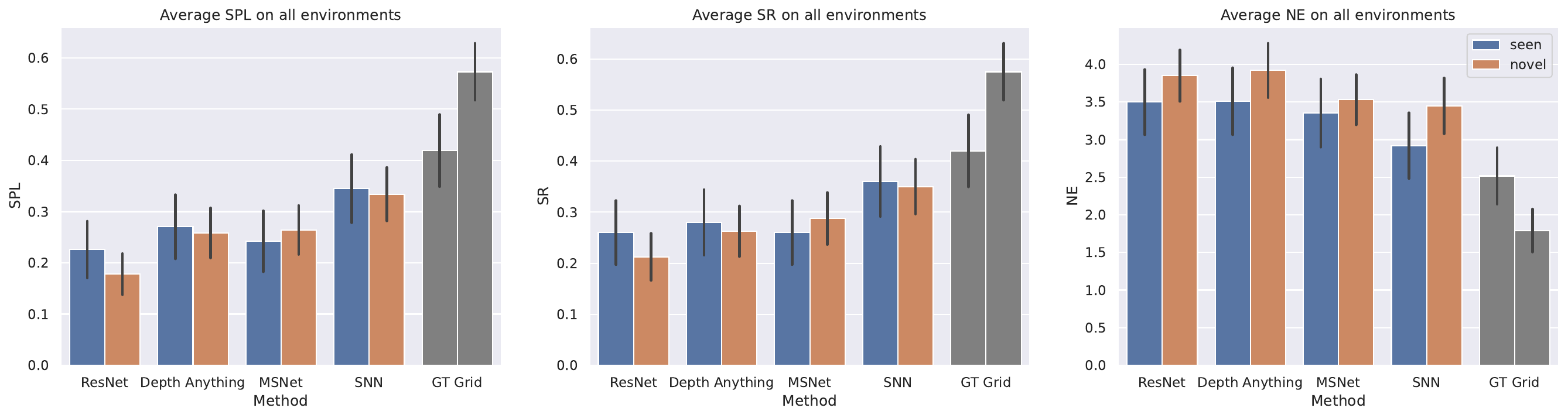}
    \caption{\textbf{Comparison against baseline approaches.} We compare against agents using ResNet, Depth Anything, MobileStereoNet (MSNet), and ground-truth voxel occupancy grid (GT Grid) using three metrics: SPL, SR, and NE. The agent using ground truth only serves as the upper bound for our policy module and is not comparable. The error bars show the standard errors.}
    \label{fig: avg-results}
\end{figure*}

\subsection{Policy Module}
\label{sec: policy_module}
Our policy module $f_o$ takes the extracted obstacle feature $h_o$ from the perception module and a point goal feature $h_g$ as input and predicts a velocity action $\hat{a}$, such that $\hat{a} = f_o(h_o, h_g)$.
The point goal $g$ is represented in the form of $(d,\cos(\theta),\sin(\theta))$ (converted from polar coordinate~\cite{wijmans_dd-ppo_2020}), where $d$ and $\theta$ are the distance and the angle relative to the goal respectively.
Before passing the goal $g$ into the policy network, we encoded it into a feature vector $h_g \in \mathbb{R}^{16}$ using a non-linear layer.

To maintain simplicity, we implement the policy module as a simple multilayer perceptron (MLP) in accordance with prior works~\cite{gandhi_learning_2017, kahn_land_2021, akmandor_deep_2022}. 
The MLP is composed of four fully connected (FC) layers, each of which is followed by a batch normalization (BN) layer and a ReLU activation function. 
At each time, the MLP predicts a velocity pair $a = (v, \omega)$ that is executed by the robot.
We optimize the policy module using behavior cloning with the Mean Squared Error (MSE) loss $\mathcal{L}_o$,
\begin{equation}
    \mathcal{L}_o = (a - \hat{a})^2,
\end{equation}
where $a$ is the ground-truth action from the sampled trajectories in the expert demonstrations, and $\hat{a}$ is the predicted action obtained by $\hat{a} = f_o(h_o, h_g)$.

\subsection{Modular Learning}
We optimize SNN using modular learning and find this leads to better navigation performance (\cref{tab: modular}).
We first train the perception module, which produces geometric obstacle features $h_o$.
Then, we feed these geometric obstacle features along with goal features $h_g$ into the policy module to predict a velocity action $a$.
Our experiments show that this design leads to better generalizability when navigating in novel environments.
Specifically, we optimize the perception module for 150 epochs, followed by optimizing the policy module for 50 epochs. 
Finally, we conduct joint fine-tuning of both modules for 300 epochs using the overall loss, which is a weighted sum of perception loss and policy loss $\mathcal{L}=\mathcal{L}_p + \alpha \mathcal{L}_o$ and can be expanded as:
\begin{equation}
    \mathcal{L} = 1 - \frac{|\calG \cap \widehat{\calG}|}{|\calG \cup \widehat{\calG}|} + \alpha (a - \hat{a})^2 ,
\end{equation}
where $\hat{a} = f_o(f_p(I_L, I_R), h_g)$.
We empirically tune the weight $\alpha=0.1$.

\section{Experiment}

We follow the privileged learning framework~\cite{chen_learning_2020, loquercio_learning_2021, sorokin_learning_2022} and collect an expert demonstration dataset using the five environments shown at \cref{fig: train-scene}.
Concretely, the privileged agent (expert) has access to the ground-truth state of all obstacles and the global occupancy map.
We generate optimal paths using the A* algorithm and utilize the PD controller to follow the planned trajectory.
The velocity command produced by the controller at every frame is saved as the expert action.

We adopt a subset of the scenes provided by the OmniGibson framework~\cite{li_behavior-1k_2023}.
We utilize five scenes during data collection and thirteen scenes (including eight novel scenes) during testing (\cref{tab: per-scene}).
To improve the robustness of policies, we apply domain randomizations to the scenes, including dome lights with different RGB colors and intensities.
We render and capture the images at the resolution of $672 \times 376$.

For each trajectory collection, the initial position of the robot and the goal position are randomly picked in the feasible areas.
We randomly sample 100 trajectories in each of the five scenes, creating a dataset with a total of 500 trajectories.
The simulation runs at real-time speed, and the collection process typically takes less than 6 hours.
The small scale of trajectories collected makes us incomparable with reinforcement learning works like DD-PPO~\cite{wijmans_dd-ppo_2020}, which leverages 80 years of navigation experience.

After collecting the expert demonstration dataset, we train the models using behavior cloning (BC).
We implement the neural networks using PyTorch and utilize Adam \cite{kingma_adam_2017} optimizer with the learning rate at 0.001. 
We divide the collected demonstration dataset from the five scenes into 80\% / 20\% split for training and validation and utilize thirteen scenes (including eight novel scenes) for testing purposes.
We set the grid size $n_x=n_y=n_z=64$ to capture details of the scenes while maintaining low computation costs during navigation.

\begin{table*}[ht!]
\centering
\caption{\textbf{Per scene numerical results.} We compare our method (SNN) against agents using ResNet (RN), Depth Anything (DA), MobileStereoNet (MS), and GT Grid (GT). The best results are \textbf{bolded}, and the second-best results are \underline{underscored}. GT is listed only for reference but is not comparable.
}
\label{tab: per-scene}
 \begin{tabular}{ c | l || r | r | r | r | g || r | r | r | r | g || r | r | r | r | g } 
 \hline
& \multirow{2}{*}{Scene} & \multicolumn{5}{c||}{SR (\%) $\uparrow$} & \multicolumn{5}{c||}{SPL (\%) $\uparrow$} & \multicolumn{5}{c}{NE (meters) $\downarrow$} \\ 
& & RN & DA & MS & SNN & \color{black}{GT} & RN & DA & MS & SNN & \color{black}{GT} & RN & DA & MS & SNN & \color{black}{GT} \\
 \hline
\parbox[t]{2mm}{\multirow{6}{*}{\rotatebox[origin=c]{90}{Seen}}} & Beechwood\_0\_int & \textbf{0.40} & 0.00 & 0.10 & \underline{0.30} & 0.60 & \textbf{0.38} & 0.00 & 0.09 & \underline{0.30} & 0.60 & \textbf{3.10} & \underline{6.07} & 6.17 & 3.74 & 1.83 \\
& Beechwood\_1\_int & 0.00 & \underline{0.20} & \textbf{0.30} & 0.00 & 0.20 & 0.00 & \underline{0.20} & \textbf{0.29} & 0.00 & 0.20 & 7.05 & 5.08 & \textbf{3.79} & \underline{5.68} & 3.92 \\
& Benevolence\_0\_int & \textbf{0.90} & 0.70 & 0.40 & \underline{0.80} & 0.80 & \textbf{0.75} & 0.66 & 0.33 & \underline{0.72} & 0.80 & \textbf{0.24} & \underline{0.38} & 1.12 & 0.32 & 0.39 \\
& Benevolence\_1\_int & 0.00 & \textbf{0.30} & \underline{0.20} & \textbf{0.30} & 0.30 & 0.00 & \textbf{0.30} & 0.20 & \textbf{0.30} & 0.30 & 4.39 & \textbf{2.90} & \underline{4.06} & 3.16 & 3.67 \\
& Benevolence\_2\_int & 0.00 & 0.20 & \underline{0.30} & \textbf{0.40} & 0.20 & 0.00 & 0.20 & \underline{0.30} & \textbf{0.40} & 0.20 & \underline{2.71} & 3.13 & \textbf{1.62} & 1.71 & 2.77 \\
\rowcolor{Gray}
& Seen Average & 0.26 & \underline{0.28} & 0.26 & \textbf{0.36} & 0.42 & 0.23 & \underline{0.27} & 0.24 & \textbf{0.34} & 0.42 & 3.50 & 3.51 & \underline{3.35} & \textbf{2.92} & 2.52 \\
\hline
\parbox[t]{2mm}{\multirow{8}{*}{\rotatebox[origin=c]{90}{Novel}}} & Ihlen\_0\_int & \textbf{0.90} & 0.50 & 0.40 & \underline{0.60} & 1.00 & \textbf{0.76} & 0.49 & 0.36 & \underline{0.59} & 1.00 & \textbf{0.40} & 2.20 & \underline{1.60} & 1.52 & 0.00 \\
& Ihlen\_1\_int & 0.00 & 0.00 & \textbf{0.30} & \underline{0.20} & 0.40 & 0.00 & 0.00 & \textbf{0.29} & \underline{0.20} & 0.40 & 5.16 & \underline{4.63} & \textbf{3.54} & 3.74 & 2.81 \\
& Merom\_0\_int & \underline{0.50} & \textbf{0.70} & 0.40 & 0.40 & 0.80 & \underline{0.41} & \textbf{0.70} & 0.33 & 0.31 & 0.80 & 1.76 & \textbf{1.49} & 3.38 & \underline{2.31} & 0.70 \\
& Merom\_1\_int & \underline{0.30} & \underline{0.30} & \textbf{0.40} & \textbf{0.40} & 0.50 & 0.25 & 0.30 & \underline{0.36} & \textbf{0.37} & 0.49 & 2.99 & 4.62 & \textbf{2.62} & \underline{3.74} & 2.46 \\
& Pomaria\_0\_int & 0.00 & \underline{0.20} & \underline{0.20} & \textbf{0.30} & 0.60 & 0.00 & 0.19 & \underline{0.20} & \textbf{0.30} & 0.60 & 5.88 & \underline{5.37} & 4.67 & \textbf{4.29} & 1.44 \\
& Pomaria\_1\_int & 0.00 & 0.10 & \underline{0.20} & \textbf{0.40} & 0.30 & 0.00 & 0.09 & \underline{0.20} & \textbf{0.39} & 0.30 & 4.77 & \underline{4.51} & \textbf{4.05} & 4.45 & 3.65 \\
& Rs\_int & 0.00 & \underline{0.10} & \textbf{0.30} & \textbf{0.30} & 0.40 & 0.00 & \underline{0.10} & \textbf{0.30} & \textbf{0.30} & 0.40 & \underline{2.65} & \textbf{2.53} & 2.61 & 2.70 & 1.35 \\
& grocery\_store\_cafe & 0.00 & \textbf{0.20} & \underline{0.10} & \textbf{0.20} & 0.60 & 0.00 & \textbf{0.20} & \underline{0.07} & \textbf{0.20} & 0.60 & 7.20 & \underline{5.99} & 5.77 & \textbf{4.85} & 1.92 \\
\rowcolor{Gray}
& Novel Average & 0.21 & 0.26 & \underline{0.29} & \textbf{0.35} & 0.57 & 0.18 & \underline{0.26} & \underline{0.26} & \textbf{0.33} & 0.57 & 3.85 & 3.92 & \underline{3.53} & \textbf{3.45} & 1.79 \\
\hline
& All Average & 0.23 & 0.27 & \underline{0.28} & \textbf{0.35} & 0.52 & 0.20 & \underline{0.26} & \underline{0.26} & \textbf{0.34} & 0.51 & 3.72 & 3.76 & \underline{3.46} & \textbf{3.25} & 2.07 \\
\hline
\end{tabular}
\end{table*}

We adopt three metrics to measure navigation performance:
\begin{enumerate}
    \item Success Rate (SR): portion of successful trials over total runs conducted.
    \item Success weighted by Path Length (SPL)~\cite{anderson_evaluation_2018}: success rate weighted by the actual travel distance compared to optimal path length.
    \item Navigation Error (NE): the mean distance to goal when failed (0 when successful).
\end{enumerate}

\begin{figure*}[t!]\centering
\vspace*{0.15cm}
\noindent 
\begin{tabularx}{\textwidth}{c *{5}{>{\centering\arraybackslash}X}}
    \rotatebox[origin=c]{90}{Benevolence\_0\_int} &
    \raisebox{-0.5\height}{\includegraphics[width=.85\textwidth]{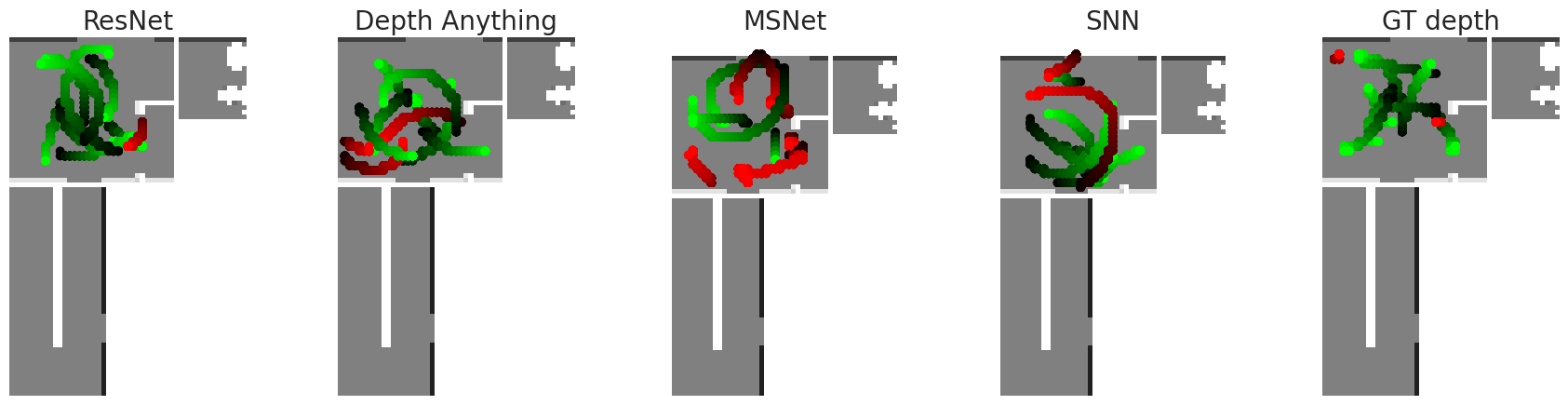}} \\

    \rotatebox[origin=c]{90}{Ihlen\_0\_int} &
    \raisebox{-0.5\height}{\includegraphics[width=.85\textwidth]{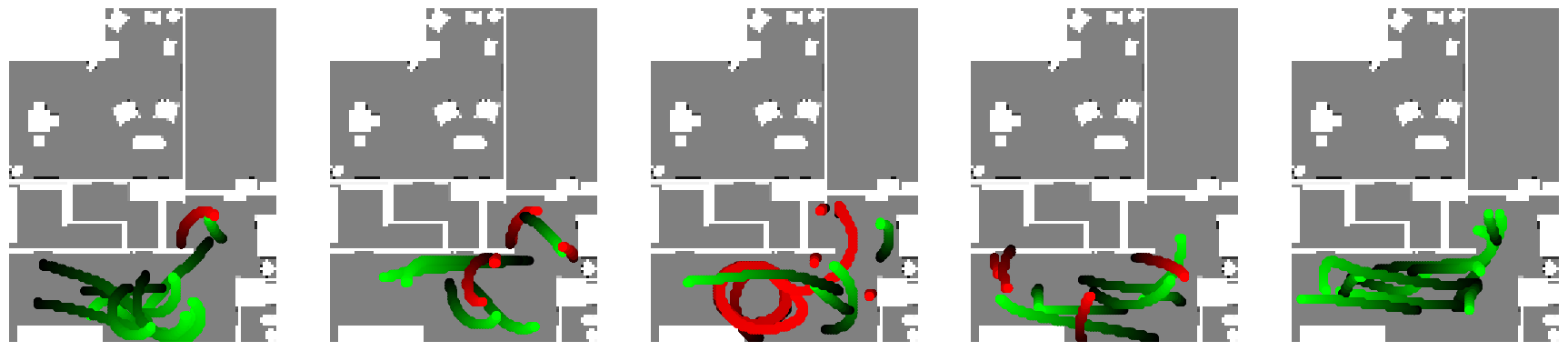}} \\
    
    \rotatebox[origin=c]{90}{Pomaria\_1\_int} &
    \raisebox{-0.5\height}{\includegraphics[width=.85\textwidth]{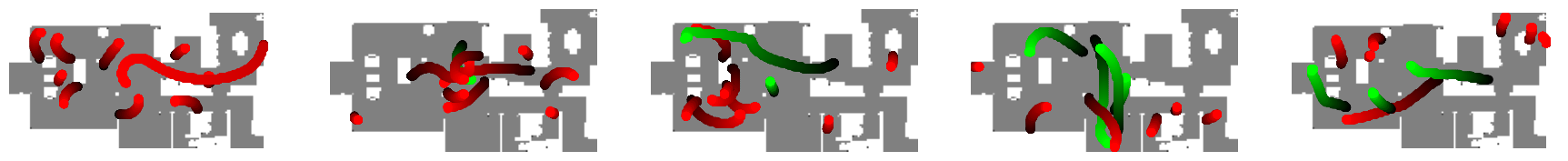}} \\

    \rotatebox[origin=c]{90}{Rs\_int} &
    \raisebox{-0.5\height}{\includegraphics[width=.85\textwidth]{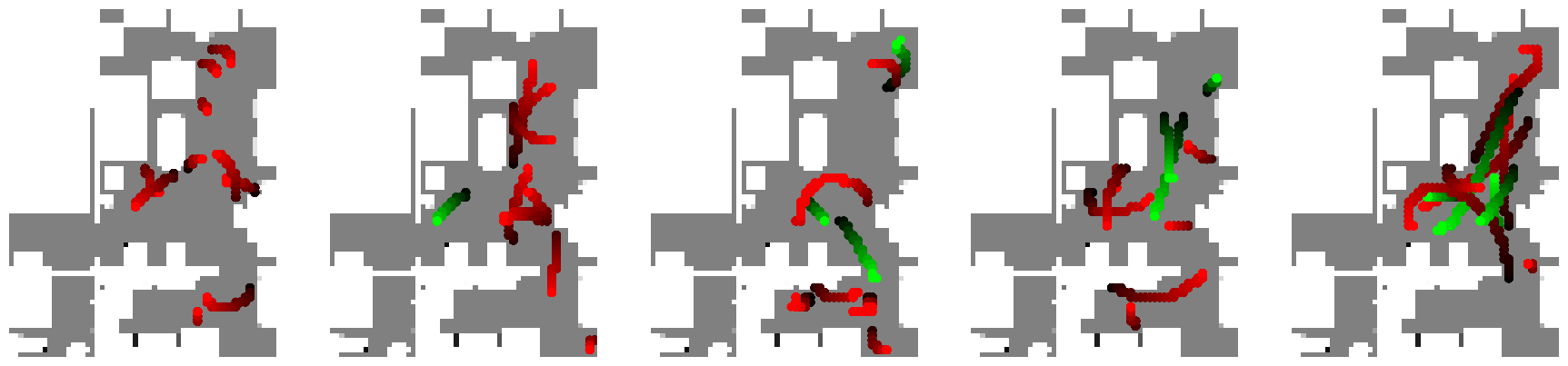}} \\
\end{tabularx}
\caption{\textbf{Qualitative results}. We present the visualizations of navigation experiments from four scenes. The top environment is seen in the demonstration dataset, and the rest are novel environments. We label the successful trials in green and the failures in red. The agents start from the light color and move towards the dark color.}
\label{fig: nav-exp}
\end{figure*}

The qualitative results of navigation performance can be found at \cref{fig: nav-exp} as well as the supplementary video.

\subsection{Comparison against baselines}
To evaluate the performance of our proposed method, we conduct experiments in five seen and eight novel environments.
To ensure a fair comparison, all of the methods listed below utilize the exact \textit{same policy module} $f_o$, and the only difference is the perception module $f_p$.
This design follows prior works such as~\cite{loquercio_dronet_2018, kareer_vinl_2023, sorokin_learning_2022}, and we only keep the FC layer and remove the LSTM layer for simplicity.
We compare our method with four approaches:
\begin{enumerate}
    \item ResNet: We leverage a ResNet visual encoder pretrained on ImageNet~\cite{russakovsky_imagenet_2015} dataset to produce a semantic obstacle feature $h_o$ from the RGB observation.
    \item Depth Anything: We first utilize the current state-of-the-art monocular depth estimation model Depth Anything~\cite{yang_depth_2024} to estimate a depth map based on the RGB image. Then we pass this depth map into the ResNet encoder to produce the geometric obstacle feature $h_o$~\cite{seo_learning_2023, wijmans_dd-ppo_2020, gervet_navigating_2023}.
    \item MobileStereoNet: Similar to Depth Anything, we leverage a stereo depth estimation module MobileStereoNet-3D~\cite{shamsafar_mobilestereonet_2022} to obtain the depth map and pass it to the ResNet encoder.
    \item GT Grid: This approach assumes access to the ground-truth depth map, which serves as a performance upper bound for our experiments. We convert the ground-truth depth map into a voxel grid $\calG$, which is the label of our perception module. 
\end{enumerate}

In the average performance shown at \cref{fig: avg-results}, our approach outperforms all of the compared approaches except GT Grid.
Notably, the GT Grid agent exhibits better performance in novel scenes compared to seen scenes. 
Two primary factors may contribute to this discrepancy:
Firstly, the GT Grid agent benefits from using ground-truth depth information, minimizing any domain gap in perception.
Secondly, the randomly selected novel scenes may pose less complexity and difficulty compared to the seen scenes.

SNN achieves SR of 35\%, SPL of 34\%, and NE of 3.25$m$, surpassing MobileStereoNet (28\%, 26\%, 3.46$m$) and Depth Anything (27\%, 26\%, 3.76$m$).
These findings suggest that the geometric feature extracted by our approach is better suited for the navigation policy.
Additionally, all hybrid approaches (SNN, Depth Anything, and MobileStereoNet) outperform the semantic counterpart (ResNet) by a large margin.

To assess generalizability, we compare the performance gaps between seen and novel scenes. 
In terms of SR and SPL, SNN experiences a drop of 1\%, while ResNet exhibits a more substantial decrease of 5\%.
This underscores the robustness of SNN in comparison to the semantic approach.

\begin{table}[t!]
\caption{\textbf{Computation cost of each approach.} We quantify the multiply-accumulate (MACs) and model parameters for each approach, alongside assessing the real-time capability by measuring its operational frequency.}
\label{tab: computation-cost}
    \centering
    \begin{tabular}{c||r|r|c}
    \hline
         Method & MACs $\downarrow$ & Params $\downarrow$ & real-time \\
    \hline
    ResNet & \textbf{1.82G} & 11.35M & \cmark \\
    Depth Anything & 91.38G & 35.52M & \xmark \\
    MobileStereoNet & 300.34G & 13.11M & \xmark \\
    SNN (Ours) & 9.53G & \textbf{1.15M} & \cmark \\
    \hline
    \end{tabular}
\end{table}

Moreover, we present the computation costs of each approach (\cref{tab: computation-cost}). 
We introduce three computation cost metrics: multiply-accumulate (MACs),  model parameters (Params), and real-time.
MAC represents the number of computational operations performed, with higher MACs typically indicating greater computational power but slower processing frequency.
Larger model parameters contribute to increased memory consumption and result in larger weight files.
For the last metric, an agent is deemed real-time if it operates at a frequency exceeding 30 Hz on an NVIDIA RTX 4070 GPU.
The results show that the other two hybrid approaches have significantly higher computation costs than SNN, resulting in the inability to operate in a real-time manner.

\subsection{Does modular learning help?}

\begin{table}[t!]
\caption{\textbf{Impact of modular learning.} We measure navigation performance and computation costs for policies built upon various grid sizes.}
\label{tab: modular}
    \centering
    \begin{tabular}{c||r|r|r}
    \hline
         Method & SR $\uparrow$ & SPL $\uparrow$ & NE $\downarrow$ \\
    \hline
         With modular learning & \textbf{0.35} & \textbf{0.34} & \textbf{3.25} \\
         Without modular learning & 0.22 & 0.22 & 3.66 \\
    \hline
    \end{tabular}
\end{table}

We leverage modular learning during our SNN training.
Concretely, we first train the perception module using only stereo data; then, we train the policy module using ground-truth perception.
Once two modules are trained to converge, we optimize SNN in an end-to-end manner.
We find this leads to higher training efficiency and stability, leading to higher ultimate navigation performance.
In this experiment, we investigate whether this paradigm improves performance.
As a comparison, we train an agent without modular learning by directly learning the outcomes of both the perception module and policy module end-to-end (same as the last step of our modular learning procedure).
As shown in \cref{tab: modular}, without modular learning, the success rate of SNN drops from 35\% to 22\% (37\% relative).
Similarly, SPL drops 35\%, and NE rises by 13\%.

We find that this is mainly due to the challenges of optimizing the perception module.
Given a randomly initialized perception module, the policy module could only obtain noisy and meaningless input.
As a result, the policy module cannot capture geometric features and further hinders the optimization of the perception module.
Beyond improved navigation performance, we envision this modular design to improve sim-to-real transfer in future studies and elaborate in the last section (\cref{sec: conclusion}).

\subsection{Does voxel grid resolution matter?}

\begin{table}[t!]
\caption{\textbf{Navigation performance with various grid sizes.} We measure navigation performance and computation costs for policies built upon various grid sizes.}
\label{tab: vary_grid}
    \centering
    \begin{tabular}{c||r|r|r|r|r}
    \hline
         Grid Size & SR $\uparrow$ & SPL $\uparrow$ & NE $\downarrow$ & MACs $\downarrow$ & Params $\downarrow$ \\
    \hline
         $64\times64\times64$ & \textbf{0.52} & \textbf{0.51} & \textbf{2.07} & 9.53G & 1.15M \\
         $16\times16\times16$ & 0.39 & 0.39 & 2.72 & 6.22G & 1.06M \\
         $4\times4\times4$ & 0.22 & 0.20 & 3.85 & \textbf{4.53G} & \textbf{0.95M} \\
    \hline
    \end{tabular}
\end{table}

In this experiment, we investigate the effect of lower voxel grid resolution on the navigation policy.
While decreasing the resolution of the voxel grid could reduce the costs of computation, especially at the perception stage, it limits the agent's understanding of the environment at lower granularity.
We train two agents with a ground-truth occupancy grid resolution of $16\times16\times16$ and $4\times4\times4$ as their input, which is 4 and 16 times smaller than our original policy trained on $64\times64\times64$ grid and present the results at \cref{tab: vary_grid}

From the results, we can observe that the navigation performance drops greatly from 52\% success rate to 39\% success rate (reduced by 25\%) when the grid size is smaller by four times.
At the same time, the computation cost (MACs) reduces by 35\%, having a slightly larger reduction than the performance.
The success rate further reduces to 22\% when using the grid of $4\times4\times4$, and SPL and NE share a similar pattern.
These findings indicate that our approach could perceive the environment in coarser-grained time-sensitive scenarios with the tradeoff of weaker navigation performance. 

\section{Conclusion}
\label{sec: conclusion}

This paper contributed SNN, a novel visual navigation network encapsulating an auxiliary voxel occupancy grid extracted from an RGB stereo camera.
We presented empirical evidence showing that SNN outperforms the semantic and hybrid baselines with better SPL, SR, and NE metrics.
We further showed the better generalizability demonstrated by SNN compared to the semantic approach and higher computation efficiency compared to other hybrid approaches.
We performed ablation studies to confirm the effectiveness of modular learning and the effect of various voxel grid sizes.
Our work shows the possibility of utilizing an auxiliary 3D representation for visual navigation that improves performance and efficiency.

\textbf{Limitations.}
In this work, we examine the generalizability of all approaches by testing them in simulated novel environments.
However, their transferability in the real world is still in question.
Nevertheless, recent works have shown great potential to achieve sim-to-real transfer using a photorealistic simulator~\cite{ehsani_imitating_2023} or modular learning~\cite{gervet_navigating_2023}, like the design we employed in this work.
Another limitation is the lack of implicit (LSTM) or explicit memory (mapping) of the environment.
These techniques could lead to better navigation performance, but they are out of the scope of this work.

\textbf{Future works.}
In future works, we are interested in investigating scaling up the perception module.
Thanks to the modular design, our perception module could be pre-trained on large-scale real-world stereo datasets.
These datasets are cheaper and have a larger scale than those obtained through robot navigation in the real world and potentially lead to policies with a narrower sim-to-real gap~\cite {li2023stereovoxelnet}.
We are also interested in achieving planning capability using differentiable planners~\cite{zhao2e2}.

\footnotesize{
\bibliographystyle{IEEEtranN}
\bibliography{zotero, custom}
}
\end{document}